\documentclass{article}

\usepackage[dblblindworkshop,final]{neurips_2026}

\workshoptitle{Foundation Models for the Brain and Body}

\usepackage[utf8]{inputenc} 
\usepackage[T1]{fontenc}    
\usepackage{hyperref}       
\usepackage{url}            
\usepackage{booktabs}       
\usepackage{amsfonts}       
\usepackage{nicefrac}       
\usepackage{microtype}      
\usepackage{xcolor}         

\usepackage{graphicx}
\usepackage{subcaption}

\title{Decomposing Brain-LLM Alignment with Interpretable Semantic Representations}

\author{Camila Kolling \qquad Mariya Toneva\\
Max Planck Institute for Software Systems (MPI-SWS) \\
\texttt{\{ckolling,  mtoneva\}@mpi-sws.org} \\
}

\begin{document}

\maketitle

\begin{abstract}
Large language models (LLMs) predict neural responses during naturalistic language comprehension, but the linguistic information underlying this brain alignment remains difficult to interpret.
We use SCRIBE, an interpretable contextual semantic representation spanning $65$ neurocognitively grounded semantic properties, as a lens for decomposing the brain alignment of one LLM, LLaMA2-7B.
Using variance partitioning between LLaMA2-7B and SCRIBE on naturalistic fMRI data from seven participants listening to narrative stories (Moth Radio Hour), we separate brain-predictive variance shared between the two representations from variance unique to each.
We find that shared variance is the dominant component across the whole brain and language-related regions, while both representations also retain significant unique contributions.
We then examine whether the concreteness of individual semantic properties relates to shared and SCRIBE-unique brain-predictive variance.
Concrete properties show greater shared variance with LLaMA2-7B than abstract properties across the whole brain and most examined regions, whereas abstract properties do not show greater SCRIBE-unique variance.
Finally, property ablations reveal that semantic contributions extend across multiple dimensions, with cumulative effects on both shared and SCRIBE-unique variance.
Together, these results show that, for LLaMA2-7B on this dataset, a substantial portion of brain--LLM alignment is associated with interpretable semantic information, while revealing complementary brain-predictive semantic information beyond the LLaMA2-7B representation.
\end{abstract}

\section{Introduction}
\label{sec:intro}

Large language models (LLMs) provide powerful models of language and have been shown to predict neural responses during naturalistic language comprehension~\cite{toneva2019interpreting, pereira2018toward}.
Yet, what linguistic information drives this brain-LLM alignment remains unclear~\cite{oota2024speech, kolling2026scribe}.
Because LLM representations jointly encode semantic, syntactic, and other linguistic information, brain predictivity alone does not establish which aspects of language underlie this alignment.

Interpretable semantic representations provide a way to isolate and characterize its semantic component.
Rather than representing meaning as a high-dimensional distributed embedding, feature-based approaches describe concepts along explicit semantic dimensions grounded in cognitive and neurobiological theories.
SCRIBE~\cite{kolling2026scribe} extends this approach to naturalistic language: it uses LLMs as semantic annotators to map words in context onto a fixed set of $65$ interpretable semantic properties spanning perceptual, motor, spatial, temporal, social, cognitive, and affective dimensions. 
This provides an explicit semantic basis against which the brain-predictive information captured by LLMs can be compared.

Here, we use SCRIBE to decompose and characterize the semantic information underlying the brain alignment of LLaMA2-7B on the Moth Radio Hour fMRI dataset~\cite{lebel2023natural}.
We perform variance partitioning between LLaMA2-7B and SCRIBE to separate brain-predictive information that is shared between the two representations from information unique to each. 
We then characterize these components at the level of semantic properties.
Specifically, we (1) quantify how much of LLaMA2-7B's brain alignment is shared with interpretable semantics, (2) test whether concrete and abstract semantic properties differ in their shared and SCRIBE-unique contributions, and (3) estimate, through systematic ablations, the unique contribution of each property given all others, as well as the contribution of cumulative sets of properties.
We focus on one LLM and one dataset for comparability with prior work (limitations in Section~\ref{sec:conclusion}).
Together, these analyses move beyond measuring whether LLMs align with the brain toward identifying what interpretable semantic information underlies that alignment.

\section{Related Work}
\label{sec:related_work}

\textbf{Brain alignment of language models.}
Contextual language-model representations predict neural responses during language comprehension~\cite{toneva2019interpreting,pereira2018toward,schrimpf2021neural}, with brain predictivity generally increasing with model scale and performance~\cite{antonello2023scaling}.
Recent work has moved beyond overall predictivity toward identifying the linguistic information underlying this correspondence, including the contribution of syntactic information to brain--LLM alignment~\cite{oota2024speech}.

\textbf{Interpretable semantic representations.}
Feature-based semantic models represent meaning through explicit, cognitively grounded dimensions and have been used to characterize semantic organization and predict neural responses~\cite{binder2016toward,mitchell2008predicting,chang2011quantitative}.
SCRIBE~\cite{kolling2026scribe} extends this framework to naturalistic language by using LLMs as contextual semantic annotators, producing interpretable semantic representations that predict brain activity beyond existing interpretable stimulus features.

Our work connects these two lines of research by directly decomposing the neural variance captured by LLM and interpretable semantic representations.

\section{Methodology}
\label{sec:method}

\subsection{Representations}

\textbf{Interpretable semantic representations.}
We use SCRIBE~\cite{kolling2026scribe} to obtain contextual and interpretable semantic representations of the language stimuli.
SCRIBE represents each content-bearing word along $65$ neurocognitively grounded semantic properties spanning sensory, motor, spatial, temporal, social, cognitive, and affective domains~\cite{binder2016toward}.
For each word, semantic property scores are estimated conditioned on its linguistic context and averaged across an ensemble of four LLM annotators (Llama-3.1-8B-Instruct, Llama-3.1-70B-Instruct, Mistral-7B-Instruct, and Qwen2.5-7B-Instruct).
SCRIBE scores have been validated against contextual semantic judgments from $921$ human participants, reaching $75$--$81\%$ of the estimated human noise ceiling~\cite{kolling2026scribe}.
We use the best-performing configuration reported in~\cite{kolling2026scribe}, considering all semantic part-of-speech categories and a context window of $k=10$ tokens.

\textbf{LLM representations.}
We use contextual representations from LLaMA2-7B~\cite{touvron2023llama}, extracted following the feature extraction procedure of Antonello et al.~\cite{antonello2023scaling}; we chose this model for direct comparability with this established procedure and with the LLaMA2-7B baselines used to evaluate SCRIBE~\cite{kolling2026scribe}.
We use layer $15$ of $32$, an intermediate layer, because intermediate LLM layers are typically among the most brain-predictive~\cite{toneva2019interpreting,antonello2023scaling}.
For each stimulus word, we extract the contextualized hidden state given the preceding words of the story, using the standard sliding context window of this procedure ($256$ to $512$ preceding words), and temporally align the resulting feature sequence with the fMRI responses.
LLaMA2-7B thus has access to a much longer context than SCRIBE's $k=10$-token window; we keep each representation in its standard configuration, so variance unique to LLaMA2-7B may partly reflect longer-range context (Section~\ref{sec:results_vp}); matching the context lengths would isolate this contribution.

\subsection{Brain encoding}
\label{sec:brain_encoding}

We evaluate brain alignment using the naturalistic fMRI dataset from the Moth Radio Hour~\cite{lebel2023natural}, in which seven participants listened to autobiographical stories while undergoing fMRI.
Word-level representations are temporally aligned with the fMRI signal using Lanczos interpolation and expanded with four delayed copies (1--4 TRs, approximately 2--8\,s) to account for the hemodynamic response.
For each voxel, we fit a ridge-regression encoding model using nine stories for training and evaluate predictions on a held-out story using Pearson correlation between predicted and observed responses.
The ridge regularization parameter is selected per voxel through bootstrapped cross-validation on the training stories; in the property ablations, the splits are fixed across full and reduced models.
Analyses are restricted to voxels with an estimated noise ceiling above $0.4$.
We report results across the whole brain and within predefined language-related ROIs, including the late language network, angular gyrus, lateral temporal cortex (LTC), and IFG/MFG.

\subsection{Variance partitioning}
\label{sec:variance_partitioning}

To characterize the semantic information underlying brain--LLM alignment, we perform voxel-wise variance partitioning between LLaMA2-7B and SCRIBE representations.
We fit three ridge-regression encoding models using LLaMA representations ($L$), SCRIBE representations ($S$), and their concatenation ($L+S$).
Let $r_L$, $r_S$, and $r_{L+S}$ denote their held-out prediction correlations.
We square these correlations and partition the resulting explained-variance estimates into variance unique to LLaMA, $U_L = r_{L+S}^2 - r_S^2$; unique to SCRIBE, $U_S = r_{L+S}^2 - r_L^2$; and shared between the representations, $C = r_L^2 + r_S^2 - r_{L+S}^2$.

The shared component $C$ quantifies brain-predictive variance captured by both representations, whereas $U_L$ and $U_S$ quantify the additional variance captured by each after accounting for the other.
Significance of each variance component is assessed using a block permutation test with $1{,}000$ permutations, disrupting prediction--response alignment while preserving local temporal structure; variance partitioning is recomputed for each permutation to form a null distribution against which the observed statistic is compared.
Comparisons across participants use paired, one-sided exact Wilcoxon signed-rank tests; with seven participants, the smallest attainable $p$-value is $1/2^7 \approx 0.0078$, reached only when all seven show an effect in the same direction, so we also report how many participants show each effect.

\subsubsection{Semantic property analysis}
\label{sec:semantic_property_analysis}

First, we use the concreteness norms of Brysbaert et al.~\cite{brysbaert2014concreteness}, which rate approximately $40,000$ English expressions on a five-point abstract--concrete scale.
Ratings are available for $61$ of the $65$ SCRIBE properties; we select the $10$ most concrete and $10$ most abstract properties (Table~\ref{tab:concreteness_properties}, Appendix~\ref{app:concreteness}) and fit separate SCRIBE and LLaMA2-7B+SCRIBE encoding models for each set, comparing their shared and SCRIBE-unique variance across participants.

Second, we perform leave-one-property-out ablations across all $65$ dimensions, retraining the SCRIBE and joint models after removing each property and ranking properties by their mean whole-brain decrease in shared and SCRIBE-unique variance.
Because SCRIBE properties are correlated (Appendix~\ref{app:collinearity}), a leave-one-out decrease measures a property's unique contribution given the other $64$, not its overall importance.
Both rankings are listed in Appendix~\ref{app:loo_rankings}.
From each ranking, we construct nested top-$5$, top-$10$, and top-$20$ sets, yielding six ablation conditions.
For each condition, we train reduced SCRIBE and joint encoding models after removing the corresponding properties from the full $65$-dimensional representation, recompute the variance partitioning, and compare the resulting shared and SCRIBE-unique components with the full model.
Because $U_S$ is small (Section~\ref{sec:results_vp}), its percentage changes are unstable; we therefore rank and test on absolute decreases and summarize $U_S$ changes by the median percentage across participants.

\section{Results}
\label{sec:results}

\subsection{Brain--LLM alignment is largely shared with interpretable semantics}
\label{sec:results_vp}

\begin{figure}[t]
    \centering
    \includegraphics[width=0.95\linewidth]{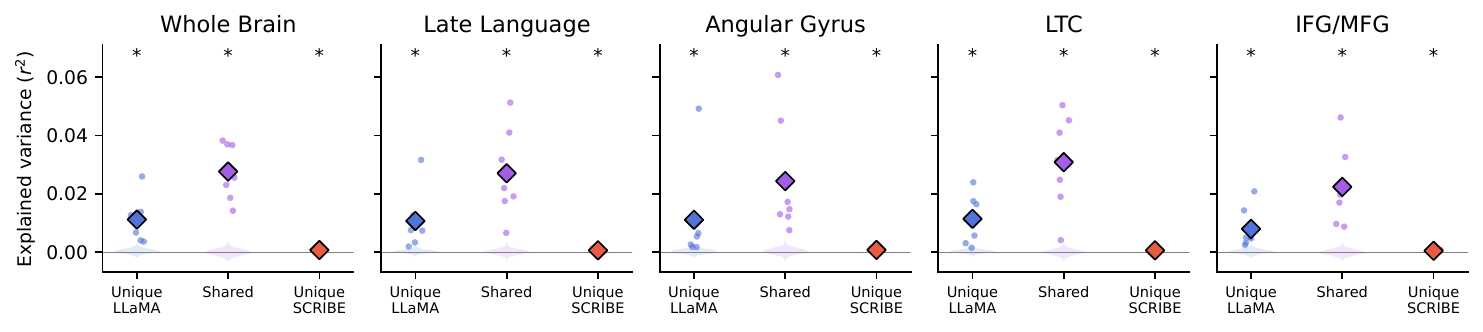}
    \caption{
    Variance partitioning between LLaMA2-7B and SCRIBE across the whole brain and language-related ROIs.
    Points show participants, diamonds show group means, and shaded distributions indicate permutation nulls.
    Explained variance (squared held-out correlation) is partitioned into components unique to LLaMA2-7B, shared between LLaMA2-7B and SCRIBE, and unique to SCRIBE.
    Asterisks indicate significance relative to the permutation distribution ($p<0.05$).
    }
    \label{fig:variance_partitioning}
\end{figure}

We first asked how much of the brain-predictive information captured by LLaMA2-7B is also captured by an explicitly semantic representation.
Variance partitioning revealed a clear pattern across the whole brain and all language-related ROIs: the variance shared between LLaMA2-7B and SCRIBE was the dominant component and was significantly above the permutation baseline in every region (Figure~\ref{fig:variance_partitioning}).
Thus, a substantial portion of LLaMA2-7B--brain alignment overlaps with information represented along SCRIBE's interpretable semantic dimensions: in the whole brain, LLaMA2-7B and SCRIBE alone explain $r^2=0.039$ and $0.028$ (joint model: $0.039$), and the shared component ($0.028$) corresponds to $71\%$ of the variance explained by LLaMA2-7B and $98\%$ of that explained by SCRIBE ($69$--$74\%$ and $97$--$98\%$ across ROIs).

The overlap was not complete: LLaMA2-7B retained significant unique variance across all regions (whole brain: $r^2=0.011$, $29\%$ of its explained variance), which may partly reflect its longer context; conversely, SCRIBE's much shorter $10$-token context did not prevent nearly all of its explained variance from being shared with LLaMA2-7B.
SCRIBE, in turn, retained a small but significant unique component (whole brain: $r^2=0.0007$) that was positive in all seven participants and exceeded the 95th percentile of its permutation null in every region, indicating a consistent rather than noisy contribution.

\begin{figure}[t]
    \centering

    \begin{subfigure}{\linewidth}
        \centering
        \includegraphics[width=0.68\linewidth]{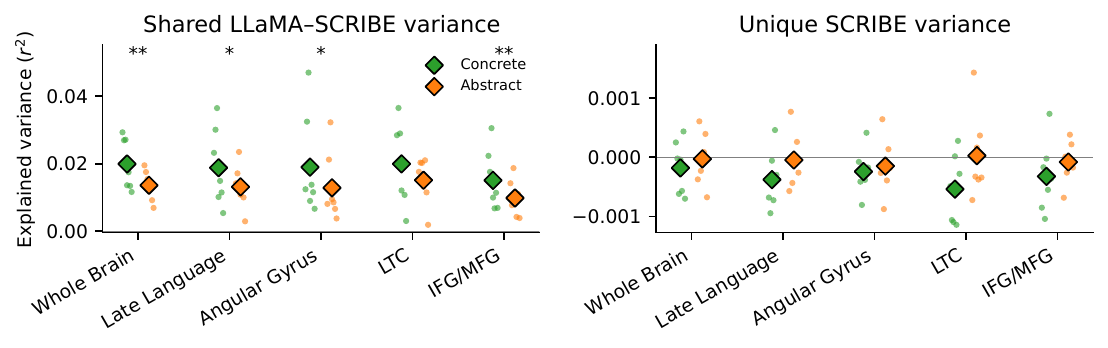}
        \caption{Concrete vs.\ abstract semantic properties.}
        \label{fig:concrete_abstract}
    \end{subfigure}

    \vspace{0.3em}

    \begin{subfigure}{\linewidth}
        \centering
        \includegraphics[width=0.68\linewidth]{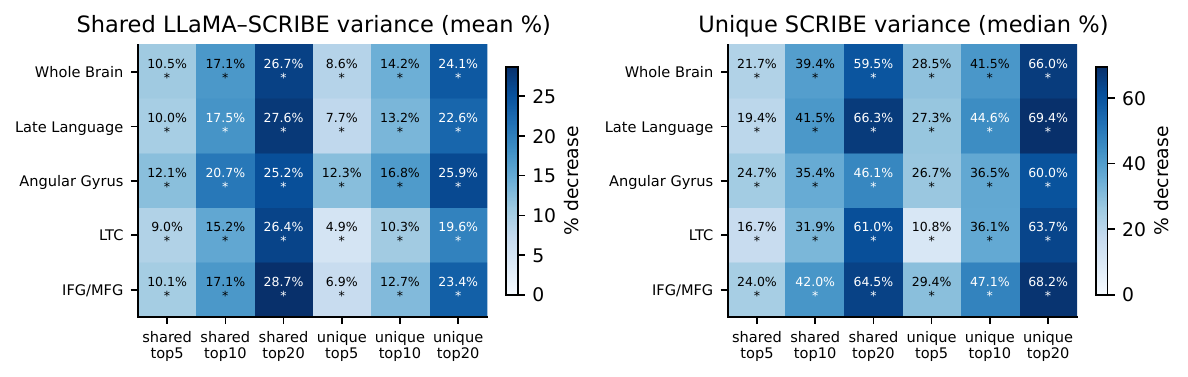}
        \caption{Cumulative semantic-property ablations.}
        \label{fig:cumulative_ablation}
    \end{subfigure}

    \caption{
    \textbf{(a)} Shared LLaMA2-7B--SCRIBE variance is greater for concrete than abstract properties in all regions except LTC, whereas SCRIBE-unique variance does not differ systematically between the two groups.
    \textbf{(b)} Percentage decrease in shared (mean) and SCRIBE-unique variance (median) after removing the top $5$, $10$, or $20$ properties ranked by their whole-brain leave-one-property-out effect on shared (left three columns) or SCRIBE-unique variance (right three columns).
    Asterisks indicate significant paired Wilcoxon signed-rank tests across participants.
    }
    \label{fig:semantic_properties}
\end{figure}

\subsection{Concrete semantics are more strongly shared with LLaMA}
\label{sec:results_concreteness}

We next asked whether shared information differs with property concreteness, comparing representations built from the $10$ most concrete and $10$ most abstract SCRIBE properties (Figure~\ref{fig:semantic_properties}a).
A consistent difference emerged for shared variance: concrete properties yielded greater shared LLaMA2-7B--SCRIBE variance than abstract properties across the whole brain and in three of the four language-related ROIs ($p\leq 0.016$; $6$--$7$ of $7$ participants), with LTC showing the same direction without reaching significance ($p=0.19$; $4$ of $7$ participants).

This distinction did not extend to the information captured uniquely by SCRIBE.
Contrary to our hypothesis, abstract properties did not explain significantly greater SCRIBE-unique variance than concrete properties in any region (all $p\geq 0.08$; Figure~\ref{fig:semantic_properties}a).
Thus, concreteness distinguishes the semantic information SCRIBE shares with LLaMA2-7B, but not the semantic information SCRIBE contributes uniquely beyond it.

\subsection{Semantic contributions extend across multiple properties}
\label{sec:results_ablation}

Individual properties had small unique effects: the largest whole-brain leave-one-out decrease in shared variance (\textit{Number}) was about $3\%$ of the shared component, and only $7$ properties decreased it in all participants (Appendix~\ref{app:loo_rankings}), as expected given the correlations among SCRIBE properties (mean $|r|=0.23$, median variance inflation factor $5.3$; Appendix~\ref{app:collinearity}): when one property is removed, part of its information remains available through correlated properties.
We therefore used the rankings to select property sets rather than to estimate individual importance, and tested whether the properties identified by the leave-one-out analysis have cumulative effects when removed together, removing the top $5$, $10$, and $20$ properties from the whole-brain shared and SCRIBE-unique rankings (Figure~\ref{fig:semantic_properties}b).

Properties selected from the shared ranking produced a clear cumulative reduction in shared LLaMA2-7B--SCRIBE variance: the decrease increased from the top-$5$ to the top-$20$ ablations across all regions, reaching approximately $25$--$29\%$ for the top-$20$ set.
In the whole brain, these decreases exceeded the sum of the individual leave-one-out decreases (top-$20$: $27\%$ vs.\ $17\%$), consistent with correlated properties carrying redundant information that is lost only when removed together.
Thus, properties with individually small effects collectively have a substantial impact on shared semantic variance when removed together.

The SCRIBE-unique component showed a similar cumulative pattern: removing properties selected from the unique ranking reduced unique variance increasingly as more properties were removed (whole-brain median $28$--$66\%$).
Moreover, removing properties selected from the shared ranking also substantially reduced SCRIBE-unique variance (median $22$--$60\%$), consistent with the large overlap between the two rankings ($14$ of the top $20$ properties; Appendix~\ref{app:loo_rankings}).
Every cumulative ablation significantly reduced both components in every region ($p\leq 0.016$; all seven participants in $57$ of $60$ tests), so SCRIBE-unique effects are consistent despite the small size of this component.
Together, both components show cumulative effects that are driven largely by a common set of semantic properties.

\section{Conclusion}
\label{sec:conclusion}

We used SCRIBE to dissect the semantic information underlying LLaMA2-7B--brain alignment in the Moth Radio Hour dataset.
A substantial component of brain-predictive information is shared with interpretable semantics, while both representations retain complementary unique contributions; this shared information was greater for concrete than abstract properties across most regions, whereas concreteness did not explain SCRIBE-unique variance.
Property ablations reveal contributions from multiple semantic dimensions, with cumulative effects for both shared and SCRIBE-unique variance.
Together, these findings show how interpretable semantic representations can move beyond aggregate brain-prediction scores to characterize the information underlying brain--LLM alignment.

\textbf{Limitations.}
We study one LLM (LLaMA2-7B, one intermediate layer), one fMRI dataset, and seven participants; our conclusions are specific to this setting, and sensitivity to the choice of layer and generalization to newer or larger LLMs, whose brain alignment varies with scale~\cite{antonello2023scaling}, remain untested.
Two of SCRIBE's annotators (Llama-3.1-8B and -70B) share LLaMA2-7B's model family; SCRIBE's validation against human judgments (Section~\ref{sec:method}) indicates that the shared component reflects human-aligned semantics, though an LLM-specific contribution cannot be fully excluded.

\clearpage

\section*{Acknowledgments}
\label{sec:acknowledgements}

This work was partially funded by the German Research Foundation (DFG) - DFG Research Unit FOR 5368 and by the Max Planck Institute for Software Systems graduate center.

\bibliographystyle{plain}
\bibliography{neurips_2026}

\appendix
\clearpage
\section{Concrete and abstract properties}
\label{app:concreteness}

\begin{table}[!h]
\centering
\caption{The 10 most concrete and 10 most abstract SCRIBE semantic properties according to the concreteness norms of Brysbaert et al.~\cite{brysbaert2014concreteness}. Higher scores indicate greater concreteness.}
\label{tab:concreteness_properties}
\small
\begin{tabular}{lc|lc}
\toprule
\multicolumn{2}{c|}{\textbf{Concrete}} &
\multicolumn{2}{c}{\textbf{Abstract}} \\
\textbf{Property} & \textbf{Score} &
\textbf{Property} & \textbf{Score} \\
\midrule
Human       & 4.93 & Pleasant       & 1.55 \\
Face        & 4.87 & Cognition      & 1.79 \\
Body        & 4.79 & Consequential & 1.86 \\
Head        & 4.75 & Complexity     & 1.87 \\
Path        & 4.41 & Unpleasant     & 1.91 \\
Music       & 4.31 & Duration       & 2.08 \\
Dark        & 4.29 & Fearful        & 2.20 \\
Landmark    & 4.10 & Away           & 2.23 \\
Color       & 4.08 & Benefit        & 2.25 \\
Taste       & 4.07 & Needs          & 2.25 \\
\bottomrule
\end{tabular}
\end{table}

\section{Correlations between SCRIBE properties}
\label{app:collinearity}

\begin{figure}[!h]
    \centering
    \includegraphics[width=0.85\linewidth]{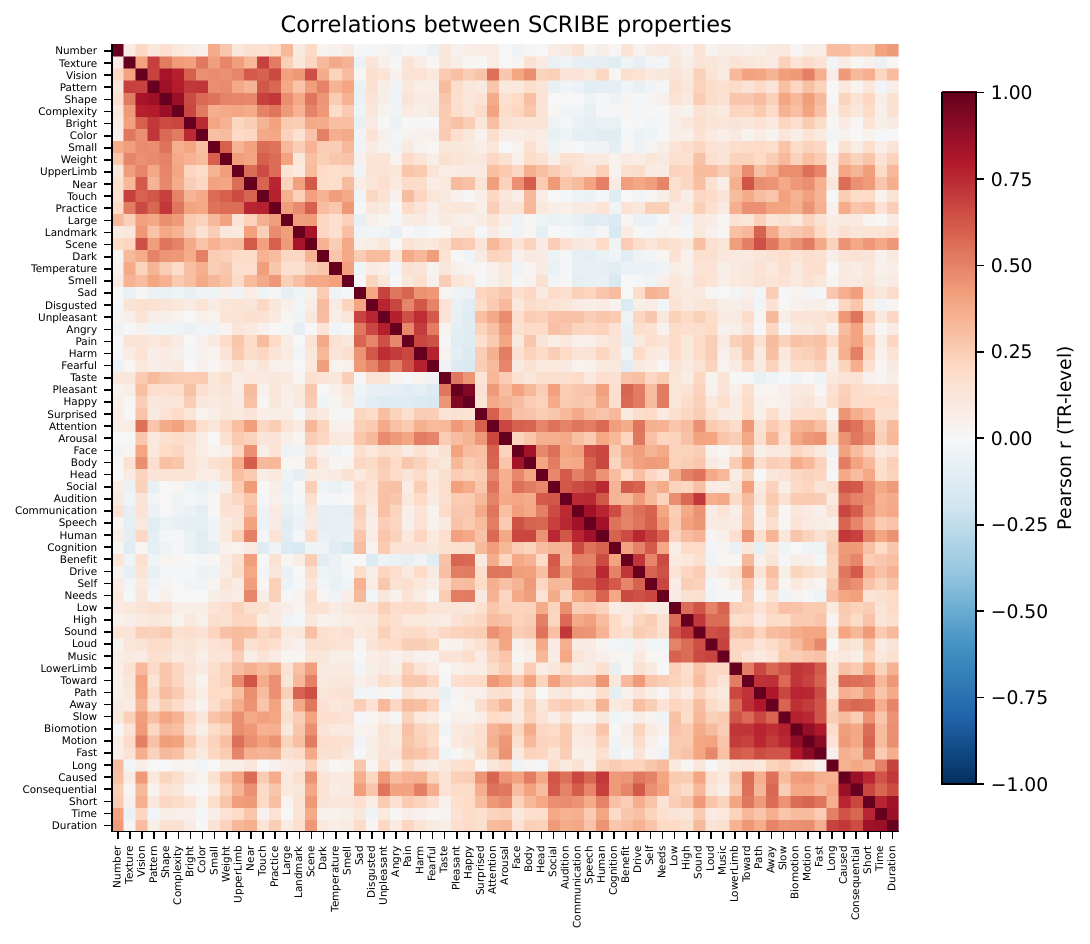}
    \caption{Pearson correlations between SCRIBE properties at the TR level, ordered by hierarchical clustering (average linkage on $1-r$).}
    \label{fig:property_correlations}
\end{figure}

Figure~\ref{fig:property_correlations} shows Pearson correlations between the $65$ SCRIBE properties, computed on the TR-level features that enter the encoding models (all stories).
The mean absolute correlation between properties is $0.23$ (median $0.18$); $10\%$ of property pairs have $|r|>0.5$ and $0.7\%$ have $|r|>0.8$, and the median variance inflation factor is $5.3$ (maximum $15.2$, \textit{Human}; seven properties exceed $10$).
Properties form clusters of related dimensions, for example visual form (\textit{Vision, Pattern, Shape, Complexity}), negative affect (\textit{Unpleasant, Angry, Harm, Fearful}), social communication (\textit{Social, Communication, Speech, Human}), motion (\textit{Path, Biomotion, Motion, Fast}), and time (\textit{Short, Time, Duration}).
Several properties at the top of the leave-one-out rankings have highly correlated partners (e.g., \textit{Landmark}--\textit{Scene}, $r=0.82$; \textit{Speech}--\textit{Human}, $r=0.84$).
This redundancy is why leave-one-property-out decreases (Section~\ref{sec:semantic_property_analysis}) are small and should be read as unique contributions given the remaining properties: removing a property whose information is shared with correlated properties changes little, even if the property is brain-predictive on its own.

\begin{figure}[h]
    \centering
    \includegraphics[width=\linewidth]{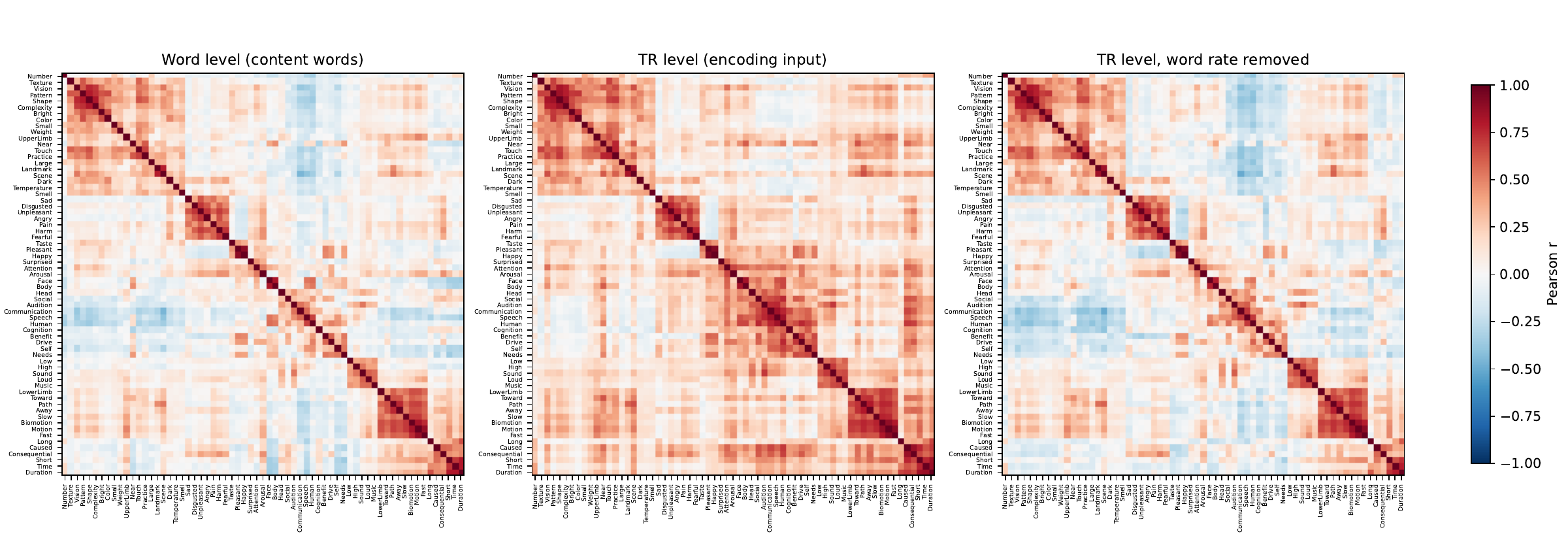}
    \caption{Correlations between SCRIBE properties across content words (left), across TRs (middle, as in Figure~\ref{fig:property_correlations}), and across TRs after regressing out word rate (right), all in the same order.}
    \label{fig:property_correlation_levels}
\end{figure}

Because the encoding models use word-level scores summed within each TR, TR-level correlations can also reflect how many content words fall in a TR (word rate) rather than meaning. We therefore also computed correlations across individual content words, before resampling, and at the TR level after regressing out the number of content words per TR (Table~\ref{tab:property_correlation_levels}, Figure~\ref{fig:property_correlation_levels}).
The structure is largely semantic: the word- and TR-level matrices are highly similar ($r=0.81$; $r=0.95$ after removing word rate), with the same strongest pairs (e.g., \textit{Pleasant}--\textit{Happy}, \textit{Motion}--\textit{Fast}, \textit{Landmark}--\textit{Scene}).
Word rate adds a positive shift to all pairs, raising TR-level correlations overall (mean $r=0.22$ vs.\ $0.08$ across words); the social-communication cluster is the most affected (\textit{Speech}--\textit{Human}: $0.84$ at the TR level vs.\ $0.68$ across words).

\begin{table}[h]
\centering
\caption{Correlations between SCRIBE properties at three levels: across content words (before resampling), across TRs (encoding-model input), and across TRs after regressing out word rate (number of content words per TR).}
\label{tab:property_correlation_levels}
\small
\begin{tabular}{lccc}
\toprule
 & Words & TRs & TRs, word rate removed \\
\midrule
Mean $|r|$ & $0.15$ & $0.23$ & $0.17$ \\
Mean $r$ & $0.08$ & $0.22$ & $0.07$ \\
Pairs with $|r|>0.5$ & $4\%$ & $10\%$ & $5\%$ \\
Negatively correlated pairs & $37\%$ & $11\%$ & $40\%$ \\
Median VIF (max) & $2.8$ ($8.6$) & $5.3$ ($15.2$) & $3.8$ ($12.8$) \\
Similarity to word-level matrix ($r$) & -- & $0.81$ & $0.95$ \\
\bottomrule
\end{tabular}
\end{table}

\clearpage
\section{Leave-one-property-out rankings}
\label{app:loo_rankings}

\begin{figure}[h]
    \centering
    \includegraphics[width=\linewidth]{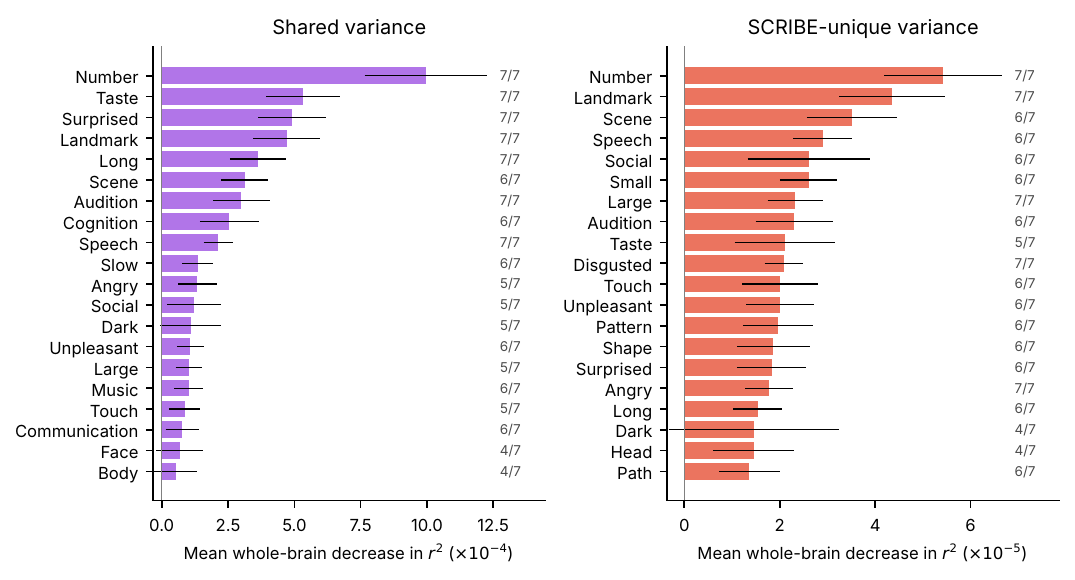}
    \caption{Top-$20$ properties by mean whole-brain decrease in shared (left) and SCRIBE-unique variance (right) when each property is removed individually. Error bars: SEM across participants; numbers: participants showing a decrease.}
    \label{fig:loo_rankings}
\end{figure}

Ranked by mean whole-brain decrease when each property is removed individually (Figure~\ref{fig:loo_rankings}), the top $20$ properties for shared variance are, in order, \textit{Number, Taste, Surprised, Landmark, Long, Scene, Audition, Cognition, Speech, Slow, Angry, Social, Dark, Unpleasant, Large, Music, Touch, Communication, Face}, and \textit{Body}; for SCRIBE-unique variance, they are \textit{Number, Landmark, Scene, Speech, Social, Small, Large, Audition, Taste, Disgusted, Touch, Unpleasant, Pattern, Shape, Surprised, Angry, Long, Dark, Head}, and \textit{Path}.

\end{document}